\documentclass{article}

\PassOptionsToPackage{numbers, compress}{natbib}
\usepackage[preprint]{optical_adversarial_attack}

\usepackage[utf8]{inputenc} 
\usepackage[T1]{fontenc}    
\usepackage{hyperref}       
\usepackage{url}            
\usepackage{booktabs}       
\usepackage{amsfonts}       
\usepackage{nicefrac}       
\usepackage{microtype}      
\usepackage{xcolor}         

\usepackage{epsfig}
\usepackage{graphicx}
\usepackage{multirow}
\usepackage{amsmath}
\usepackage{wrapfig}
\usepackage{diagbox}

\newcommand{\STAB}[1]{\begin{tabular}{@{}c@{}}#1\end{tabular}}

\title{Light Lies: Optical Adversarial Attack}

\author{
  Kyulim Kim\thanks{Department of Artificial Intelligence, Yonsei University, South Korea} \\
  \texttt{kyulim.kim@yonsei.ac.kr} \\
  \And
   JeongSoo Kim\thanks{Department of Mechanical Engineering, Yonsei University, South Korea} \\
   \texttt{wjdtn1110@yonsei.ac.kr} \\
   \And
   Seungri Song\footnotemark[2] \\
   \texttt{songsangzz@yonsei.ac.kr} \\
   \And
   Jun-Ho Choi\thanks{School of Integrated Technology, Yonsei University, South Korea} \\
   \texttt{idearibosome@yonsei.ac.kr} \\
   \And
   Chulmin Joo\footnotemark[2] \\
   \texttt{cjoo@yonsei.ac.kr} \\
   \And
   Jong-Seok Lee\footnotemark[1] \footnotemark[3] \\
   \texttt{jong-seok.lee@yonsei.ac.kr}
}

\begin{document}

\maketitle

\begin{abstract}
A significant amount of work has been done on adversarial attacks that inject imperceptible noise to images to deteriorate the image classification performance of deep models. However, most of the existing studies consider attacks in the digital (pixel) domain where an image acquired by an image sensor with sampling and quantization has been recorded. This paper, for the first time, introduces an optical adversarial attack, which physically alters the light field information arriving at the image sensor so that the classification model yields misclassification. More specifically, we modulate the phase of the light in the Fourier domain using a spatial light modulator placed in the photographic system. The operative parameters of the modulator are obtained by gradient-based optimization to maximize cross-entropy and minimize distortions. We present experiments based on both simulation and a real hardware optical system, from which the feasibility of the proposed optical attack is demonstrated. It is also verified that the proposed attack is completely different from common optical-domain distortions such as spherical aberration, defocus, and astigmatism in terms of both perturbation patterns and classification results.
\end{abstract}

\section{Introduction}
\label{sec:introduction}

It is well known that injecting small perturbations to input data can significantly degrade the performance of deep neural networks, called adversarial attacks.
Because such attacks raise security concerns of deep learning-based applications, many researchers have studied the impact of the adversarial attacks on various deep models, especially for image classification models \cite{goodfellow2014explaining,su2018robustness}.

Most existing studies focus on finding adversarial examples in the digital domain, i.e., altering the pixel values of digital images.
Another possibility is that an attack is applied to the target object in the physical domain.
For this, a few studies demonstrate the efficacy of adversarial examples found in the digital domain when they are implemented in the physical domain, e.g., printed objects \cite{kurakin2016adversarialphysical,athalye2018synthesizing}.
Such applicability of adversarial examples on real objects raises more severe security concerns in various practical applications (e.g., autonomous vehicle system \cite{nassi2019mobilbye}, person detector \cite{xu2020adversarial}).

Orthogonal to these attempts, this paper introduces an \emph{optical adversarial attack} by considering a new layer between real objects and digital images for implementing adversarial attacks, i.e., the optical system acquiring the light field information from the target object in the physical world and converting it to the image in the digital domain.
The idea is to modulate the phase of the light information in the Fourier domain using a device called spatial light modulator (SLM).
Spatially varying phase modulations are found by optimizing an objective function to minimize image distortion and maximize the cross-entropy, which are realized by the SLM in the optical system.
The change in the digital image obtained by the image sensor due to the phase modulations is hardly perceptible, but can significantly deteriorate the performance of the image classification model.

The main contribution of our work can be summarized as follows.
\begin{itemize}
	\item
	We propose an optical adversarial attack that is implemented in the optical system, which deteriorates the performance of the deep models performing classification using the images acquired from the optical system (Section~\ref{sec:system}). 
	\item
    We show the feasibility of our optical adversarial attack by conducting experiments on a simulated optical system for various images from the ImageNet dataset (Section~\ref{sec:exp_simulation}).
	It is shown that the attacked optical system produces output images that have similar quality as the original outputs but fool the subsequent image classification models.
	\item
	We conduct real experiments on an actual system implementing our attack to demonstrate the feasibility of the proposed idea in the real world (Section~\ref{sec:exp_real}).
	Our attack is also compared to common optical-domain phase distortions such as spherical aberration, defocus, and astigmatism, which verifies the significant superiority of our method as an attack. 
\end{itemize}

Our work has two important implications.
First, our work is the first to demonstrate the possibility of implementing adversarial attacks by altering the light information in the optical system.
The work in \cite{li19adversarial} proposes a physical attack by putting a sticker on the camera lens, but the attack occurs outside the optical system and, furthermore, physical intervention (i.e., putting a sticker) is required.
In contrast, our attack takes place \emph{inside} the optical system, and is implemented without physical intervention by maliciously controlling the computer used as the controller of the SLM.
Second, we raise a new immediate vulnerability issue of practical systems where SLMs are employed, including biomedical imaging, holography, and optical encryption.
In such systems, our work shows that malicious attempts may be made not only by conventional attacks in the digital domain but also by optical attacks in the physical domain.

\section{Related work}

\subsection{Adversarial attack}

Various adversarial attack methods against image classification models have been developed.
Goodfellow \emph{et al.} \cite{goodfellow2014explaining} proposed the fast gradient sign method (FGSM) that obtains a perturbation for a given image from the sign of the gradients of a target image classification model.
Kurakin \emph{et al.} \cite{kurakin2016adversarial} extended FGSM to an iterative approach to find a more powerful perturbation, which is called I-FGSM.
Carlini and Wagner \cite{carlini2017towards} developed an efficient attack method that finds a perturbation by minimizing the amount of deterioration and the distance of logits between the original predicted class label and the target label.

While the aforementioned methods focus on injecting a perturbation into a given digital image that will be directly inputted to a target image classification model, some researchers have also investigated adversarial examples that are applicable to physical objects.
Kurakin \emph{et al.} \cite{kurakin2016adversarialphysical} demonstrated the feasibility of finding adversarial examples that can fool the classification model even when the attacked images are printed and captured again using a phone camera.
Eykholt \emph{et al.} \cite{eykholt18robust} showed that physically perturbing real objects such as road signs can attack image classification models.
Athalye \emph{et al.} \cite{athalye2018synthesizing} further provided adversarial showcases with 3D-printed objects that can make the classification model misclassify the images taken in various viewpoints.

Previous research has focused on attacking images or objects themselves, and to the best of our knowledge, there is no approach that attacks optical systems acquiring images from real objects.

\subsection{SLM-based optical system}

An SLM is an computer-controlled active device used to modulate the amplitude, phase, or polarization of light waves in space and time.
Among several types of SLMs, liquid crystal on silicon (LCoS) SLMs are used in applications that call for phase modulations in optical systems such as lithography \cite{jenness2008three,jenness2010versatile,lowell2017simultaneous}, optical tweezer \cite{reicherter1999optical,kim2016situ,hadad2018particle}, turbulence simulation \cite{burger2008simulating,phillips2005atmospheric}, and imaging \cite{quirin2013instantaneous,wang2011spatial,situ2010phase,jesacher2007wavefront,warber2010combination,mukherjee2019spatial}.

In Fourier optics, a lens is regarded as a Fourier transform engine.
That is, for a given object field in the front focal plane of the lens, its Fourier transform can be obtained in the back focal plane of the lens.
This plane is referred to as the Fourier plane, where one has access to the spatial frequency spectrum of the object field.
By placing an SLM in the Fourier plane, one can alter phase delay individually for each spatial frequency component, thus modifying the transfer function or image formation of an optical imaging system.
For example, it has been shown that the depth-of-field of optical imaging systems can be increased significantly by introducing cubic phase offset in the Fourier plane \cite{quirin2013instantaneous}.
In addition, the phase modulation technology using SLMs has been used in phase imaging of thin biological specimens \cite{wang2011spatial,situ2010phase} and aberration correction of optical systems \cite{jesacher2007wavefront,warber2010combination}.
Various applications of SLMs for the pupil engineering can be referred to in the review paper \cite{maurer2011spatial}.

A recent study by Kravets \emph{et al.} \cite{kravets2021compressive} introduced a defense technique using an SLM to defend adversarial attacks applied in the digital domain.
On the other hand, we consider an optical adversarial attack, which is implemented using a phase SLM.

\section{Proposed system}
\label{sec:system}

\begin{figure}[t]
    \begin{center}
    	\centering
    	\footnotesize
    	\begin{minipage}[b]{0.99\linewidth}
    		\centering
    		\centerline{\includegraphics[width=1.0\linewidth]{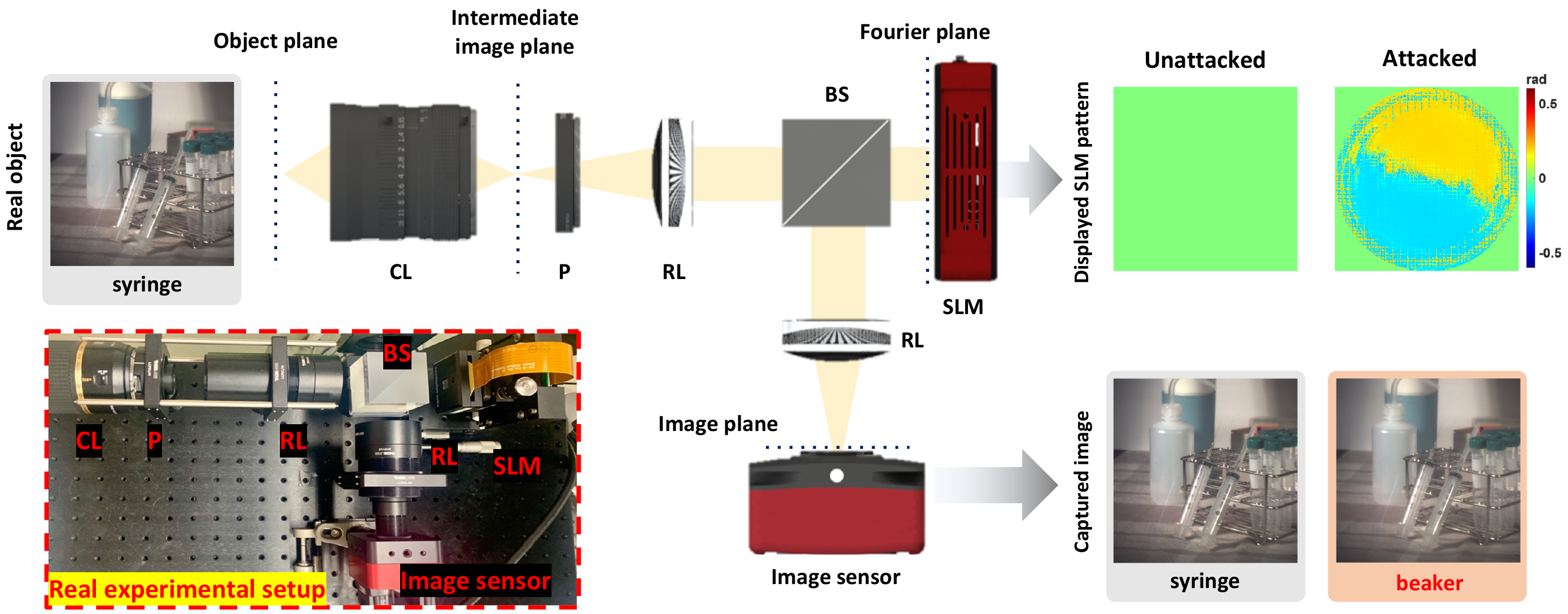}}
    	\end{minipage}
    \end{center}
    \caption{Overview of the proposed optical adversarial attack system. A phase modulation module consisting of a polarizer, relay lens, beam splitter, and SLM is implemented to the photography system. The unattacked image obtained without phase modulation and the attacked image obtained with adversarial phase modulation are acquired, respectively. When the acquired images are classified by the deep model, the unattacked image is classified correctly, but the attacked image is misclassified. CL: camera lens; P: polarizer; RL: relay lens; BS: beam splitter.}
    \label{fig:target_system_diagram}
\end{figure}

We set up an SLM-based optical system that consists of a camera lens, an SLM, and an image sensor, which is illustrated in \figurename~\ref{fig:target_system_diagram}.
The image acquisition process is as follows.
First, a commercial camera lens (SP AF 60mm F/2 Di II Macro 1:1, Tamron) collects the object field and then generates the image at the intermediate image plane.
In order to achieve direct access to the Fourier plane, we construct a 4-f system using two lenses (RL, AC508-100-A, Thorlabs) to relay the information onto the image plane.
A phase-only SLM (HSP512, Meadowlark) is placed in the Fourier plane, i.e., the back focal plane of the first relay lens.
Since the SLM is polarization-dependent, a linear polarizer (LPVISC100, Thorlabs) is placed before the SLM.
The phase-modulated light via the SLM is then reflected by the beam splitter (BS031, Thorlabs), and the image in the image plane is captured by an image sensor (Flare 4M180-CL, IO Industries).

An obtained digital image is then inputted to a deep neural network that classifies an object in the image.
In this study, we consider three widely known models, namely, ResNet50 \cite{he2016deep}, VGG16 \cite{simonyan2014very}, and MobileNetV3 \cite{howard2019searching}, which are pre-trained on the ImageNet dataset \cite{russakovsky2015imagenet}.

On this system architecture, our optical adversarial attack aims to find an adversarial perturbation that is displayed as an SLM pattern, which leads the classifier to misclassify the resulting image, while no significant visible differences are observed between the unattacked and attacked images.

\subsection{Imaging model of SLM-based optical adversarial attack}
\label{subsec:Imaging_model}

Let ${X}_{\mathrm{obj}}$ be the intensity of an object.
The camera lens forms image ${X}_{\mathrm{img}}$ at the intermediate image plane, and then the 4-f system relays this information onto the image plane (see Figure~\ref{fig:target_system_diagram}).
For an incoherent imaging system, the acquired image ${X}$ can be expressed as \cite{goodman2005introduction}
\begin{equation}
\label{eq:i_cam}
    {X} = {iPSF} \otimes {X}_{\mathrm{img}},
\end{equation}
where ${iPSF}$ denotes the incoherent point spread function and $\otimes$ represents 2D convolution. ${iPSF}$ is equal to the squared magnitude of the coherent point spread function $h$ (i.e., ${iPSF} = |h|^{2}$).
Note that $h$ is the Fourier transform of the pupil function $H$.
In the Fourier domain, (\ref{eq:i_cam}) can be written as
\begin{equation}
    \tilde {X} = {MTF} \cdot \tilde{X}_{\mathrm{img}},
\end{equation}
where ${MTF}$ is the modulation transfer function, and $\tilde{X}_{\mathrm{img}}$ is the Fourier transform of ${X}_{\mathrm{img}}$.
Using convolution theorem, ${MTF}$ can be obtained as ${MTF} = \mathcal{F} ( |h|^{2} ) = | H \star H |$, where $\mathcal{F}(\cdot)$ is the Fourier transform operator and $\star$ represents 2D correlation.

We consider a circular aperture in the pupil plane with a radius of $R$.
If a phase modulation is applied in the pupil plane, the corresponding pupil function can be expressed as
\begin{equation}
    {H} = 
    \begin{cases}
        {e}^{j \phi(\vec{u})}, & \text{if } ||\vec{u}|| < R \\
        0, & \text{otherwise}
    \end{cases}
    ,
\end{equation}
where $\vec{u}$ is the spatial frequency coordinate at the Fourier plane and $\phi(\vec{u})$ is the modulated phase distribution, which is applied through the SLM in our case.

\subsection{Finding adversarial perturbation}

In our attack, non-targeted adversarial phase perturbation $\phi(\vec{u})$ is found by a gradient-based ${l}_{2}$-norm optimization method to maximize the classification loss while minimizing image distortions.

Let $\hat{X}_{\phi}$ denote the attacked version of $X$ with phase modulation $\phi$.
The optimization problem to find $\hat{X}_{\phi}$ is written as
\begin{equation}
\label{eq:cost_function}
    \mathrm{arg}\min_{\phi} \lambda || \hat{X}_{\phi} - {X} ||_{2} - l \big( {y}, {f} (\hat{X}_{\phi}) \big),
\end{equation}
where $\lambda$ is a balancing constant between the two terms, $l$ is the classification loss (i.e., cross-entropy), $y$ is the ground truth class label, and $f$ is the classification model.
To find an appropriate value of $\lambda$, we adopt an iterative approach that starts with a large value (to ensure a small amount of image distortion) and gradually decreases it until the classification result becomes incorrect.

\section{Simulation experiments}
\label{sec:exp_simulation}

Before we apply our adversarial attack on a real system composed of physical devices, we first conduct experiments on a simulation environment using the forward model explained in Section~\ref{subsec:Imaging_model}.
This enables us to find out the feasibility of our proposed adversarial attack method by employing a relatively larger number of images containing diverse objects.

\subsection{Implementation details}

We employ 1,000 test images of the NeurIPS 2017 Adversarial Attacks and Defences Competition \cite{kurakin2018adversarial}\footnote{We obtained the images from \url{https://kaggle.com/c/6864}.}.
This dataset contains images associated with each of the 1,000 ImageNet classes, which are not included in the training images of the original ImageNet dataset.

\begin{figure}[t]
	\begin{center}
		\centering
		\footnotesize
		\begin{minipage}[b]{0.525\linewidth}
			\centering
			\centerline{\includegraphics[width=1.0\linewidth]{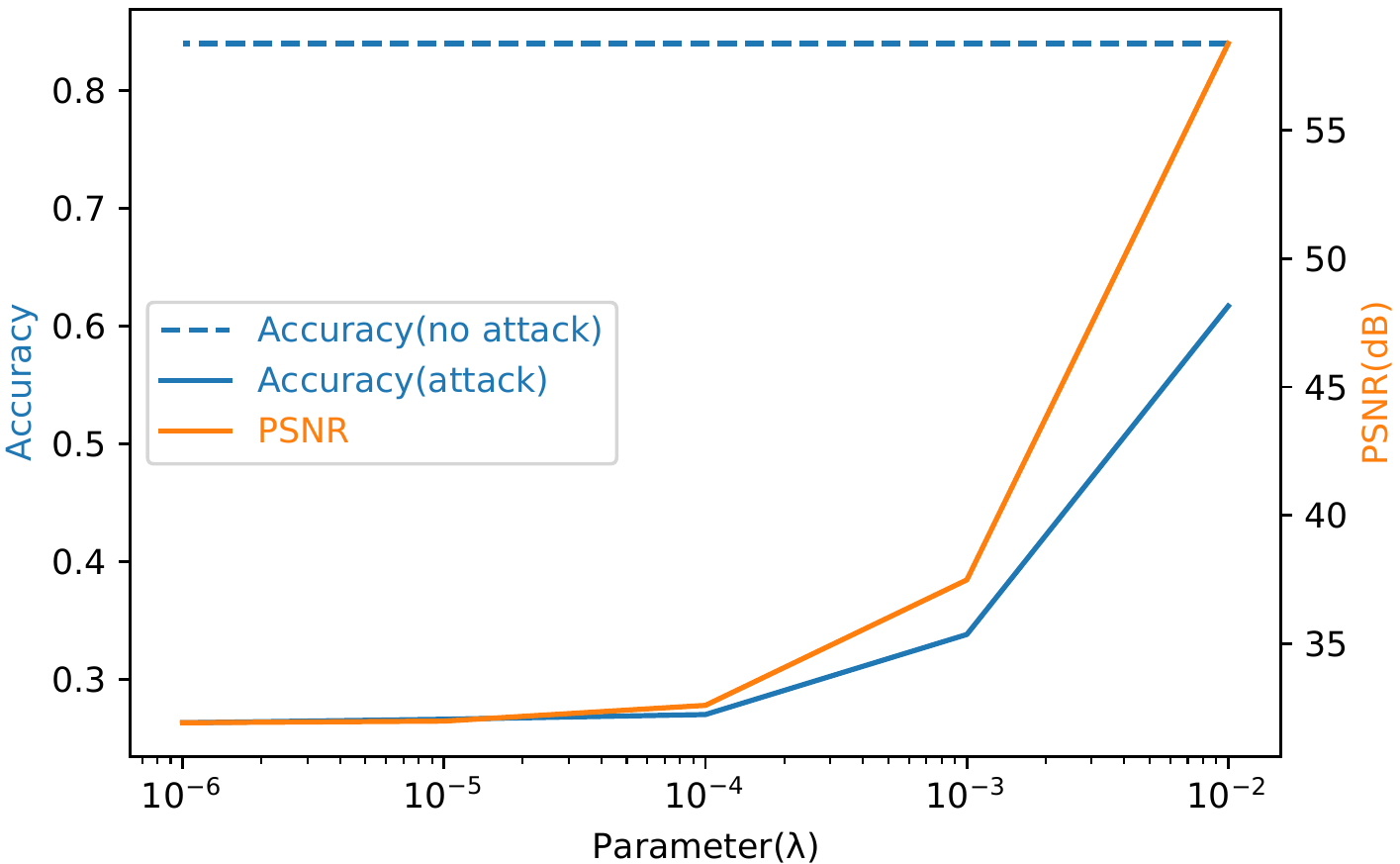}}
			\centerline{(a)}
		\end{minipage}
		~
		\begin{minipage}[b]{0.455\linewidth}
			\centering
    		\renewcommand{\tabcolsep}{1.0pt}
    		\newcommand{\spsize}{0.30\linewidth}
    		\begin{tabular}{cccc}
    		    {\scriptsize{$\lambda = 10^{-2}$}} & {\scriptsize{$\lambda = 10^{-3}$}} & {\scriptsize{$\lambda = 10^{-4}$}} & \\
    		    \includegraphics[width=\spsize]{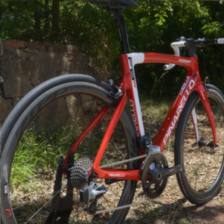} &
    		    \includegraphics[width=\spsize]{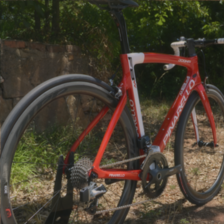} &
    		    \includegraphics[width=\spsize]{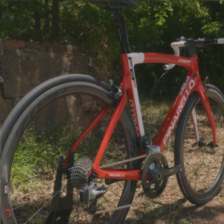} &
    		    \\
    		    \includegraphics[width=\spsize]{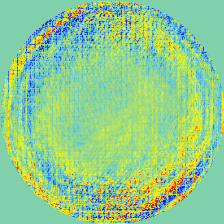} &
    		    \includegraphics[width=\spsize]{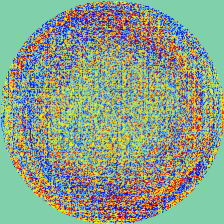} &
    		    \includegraphics[width=\spsize]{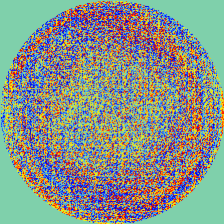} &
    		    \includegraphics[width=0.051\linewidth]{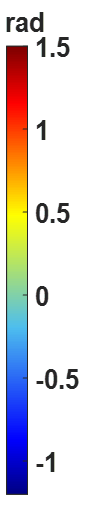} \\
    		    {\scriptsize{\texttt{mountain bike}}} & {\scriptsize{\texttt{mountain bike}}} & {\scriptsize{\color{red}{\texttt{tricycle}}}} &
		    \end{tabular}
			\centerline{(b)}
		\end{minipage}
	\end{center}
	\caption{(a) Comparison of the classification accuracy of the original and attack images with respect to the value of $\lambda$ for the VGG16 model. PSNR values calculated from the unattacked and attacked images are also shown. (b) A showcase of the attacked images (top) and their phase perturbations (bottom) for different values of $\lambda$.}
	\label{fig:acc_psnr}
\end{figure}

The classification accuracy is used as the primary evaluation metric.
In addition, we employ the peak signal-to-noise ratio (PSNR) and structural similarity (SSIM) to measure the amount of deterioration in the attacked images compared to their corresponding unattacked images.

To find an adversarial example $\hat{X}_{\phi}$ for a given image ${X}$ by (\ref{eq:cost_function}), we employ the Adam optimizer \cite{kingma2014adam} because it is known to be effective in quickly finding adversarial examples \cite{carlini2017towards}.
We use a learning rate of ${5}\times{10}^{-3}$ and a weight decay factor of ${5}\times{10}^{-6}$.
We initially set the value $\lambda$ to ${10}^{-2}$ and reduce it by $1/10$ if a valid $\hat{X}_{\phi}$ is not found within the maximum number of iterations, which is set to 150.
The optimization process stops once we obtain a valid $\hat{X}_{\phi}$ that makes the model misclassify the attacked image.

We observe that both accuracy and PSNR tend to converge to certain values as $\lambda$ decreases.
Figure~\ref{fig:acc_psnr}(a) shows such a tendency of convergence for VGG16.
When $\lambda$ becomes ${10}^{-4}$, the accuracy and PSNR are measured as 0.270 and 33.60 dB, respectively, and both do not significantly change when $\lambda$ decreases further.
Therefore, we set the minimum value of $\lambda$ as ${10}^{-6}$.
Figure~\ref{fig:acc_psnr}(b) depicts a showcase of the obtained images for different values of $\lambda$.
The three cases do not show significant perceptual differences, while the classification result becomes wrong when $\lambda$ becomes ${10}^{-4}$ and a larger amount of phase modulation is applied.

\subsection{Results}

\begin{table}[t]
  \caption{Performance comparison in terms of accuracy, PSNR, and SSIM evaluated on different image classification models. Standard deviations across images are also shown. Note that relatively large standard deviations of PSNR are due to the images with failed attack despite severe phase perturbations and the images with little changes despite the attack.}
  \label{table:simulated_results}
  \centering
  \begin{tabular}{c|cccc}
    \toprule
    Model & Accuracy (original) & Accuracy (attacked) & PSNR (dB) & SSIM \\
    \midrule
    ResNet50 & 0.896 & 0.334 & 33.98 {\scriptsize{($ \pm 9.77 $)}} & 0.9623 {\scriptsize{($ \pm 0.0495 $)}} \\
    VGG16 & 0.840 & 0.260 & 32.96 {\scriptsize{($ \pm 9.10 $)}} & 0.9561 {\scriptsize{($ \pm 0.0572 $)}} \\
    MobileNetV3 & 0.860 & 0.323 & 35.40 {\scriptsize{($ \pm 9.79 $)}} & 0.9669 {\scriptsize{($ \pm 0.0458 $)}} \\
    \bottomrule
  \end{tabular}
\end{table}

Table~\ref{table:simulated_results} shows the performance comparison on the three classification models.
When the attack method is not employed, all the models achieve classification accuracy above 0.840.
However, when our adversarial attack is employed, the accuracy values are significantly reduced.
This result proves that the optical system for the image classification task is highly vulnerable to our proposed adversarial attack.
In addition, both the PSNR and SSIM values of the images obtained from the attacked optical system are significantly high (i.e., above 30 dB).
It implies that differences between the original and attacked images are hardly noticeable.

As a baseline, we test a so-called ``random phase attack'' by constructing a random phase pattern $\phi$ that generates a digital image having a similar PSNR value to that of an image obtained from our adversarial attack.
With this method, we obtain images having an average PSNR value of 33.12 dB, which is similar to the average PSNR values in Table~\ref{table:simulated_results} and even slightly lower than those obtained from our attack for ResNet50 and MobileNetV3.
However, the classification accuracy barely drops when those images are inputted to the models, which are 0.894, 0.828, and 0.860 for ResNet50, VGG16, and MobileNetV3, respectively.
This result shows that the perturbations found by our proposed attack method are very different from random perturbations and our method successfully deteriorates the classification performance while preserving the quality of the obtained images.

\begin{figure}[t!]
	\begin{center}
		\centering
		\renewcommand{\tabcolsep}{2.0pt}
		\newcommand{\sresize}{0.125\linewidth}
		\newcommand{\srescsize}{0.023\linewidth}
		\footnotesize
		\begin{tabular}{cccccccccccc}
		    & & \multicolumn{3}{c}{Original} & \multicolumn{3}{c}{Attacked} & \multicolumn{3}{c}{Diff. (PSNR, SSIM)} & Phase
		    \\
		    \midrule
		    \multirow{3}{*}[-40pt]{\rotatebox[origin=c]{90}{ResNet50}} & &
		    \raisebox{-0.5\height}{\includegraphics[width=\sresize]{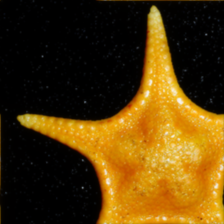}} &
		    \STAB{\texttt{starfish} \\ (70.0\%)} & &
		    \raisebox{-0.5\height}{\includegraphics[width=\sresize]{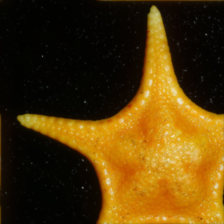}} &
		    \STAB{\texttt{\color{red}{honeycomb}} \\ (12.9\%)} & &
		    \raisebox{-0.5\height}{\includegraphics[width=\sresize]{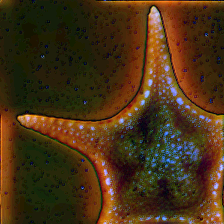}} &
		    \STAB{31.94 dB \\ 0.7712} & &
		    \raisebox{-0.5\height}{\includegraphics[width=\sresize]{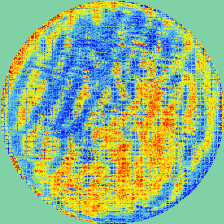} \includegraphics[width=\srescsize]{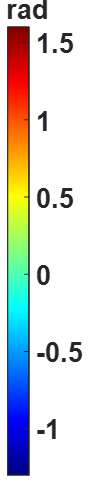}}
		    \smallskip \\
		    & &
		    \raisebox{-0.5\height}{\includegraphics[width=\sresize]{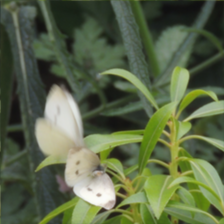}} &
		    \STAB{\texttt{cabbage} \\ \texttt{butterfly} \\ (45.1\%)} & &
		    \raisebox{-0.5\height}{\includegraphics[width=\sresize]{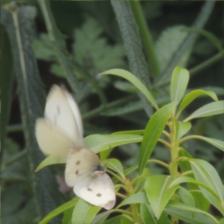}} &
		    \STAB{\texttt{\color{red}{earthstar}} \\ (22.2\%)} & &
		    \raisebox{-0.5\height}{\includegraphics[width=\sresize]{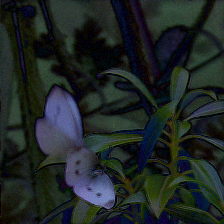}} &
		    \STAB{35.15 dB \\ 0.9930} & &
		    \raisebox{-0.5\height}{\includegraphics[width=\sresize]{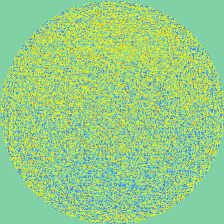} \includegraphics[width=\srescsize]{figures/simulated_results/colorbar}}
		    \smallskip \\
		    & &
		    \raisebox{-0.5\height}{\includegraphics[width=\sresize]{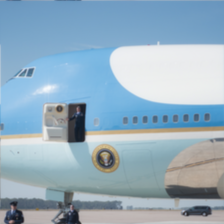}} &
		    \STAB{\texttt{airliner} \\ (94.8\%)} & &
		    \raisebox{-0.5\height}{\includegraphics[width=\sresize]{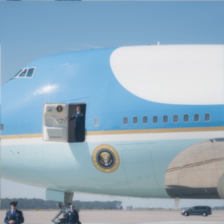}} &
		    \STAB{\texttt{\color{red}{space}} \\ \texttt{\color{red}{shuttle}} \\ (43.1\%)} & &
		    \raisebox{-0.5\height}{\includegraphics[width=\sresize]{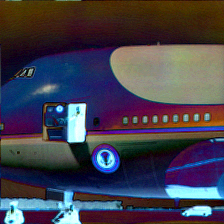}} &
		    \STAB{28.58 dB \\ 0.9840} & &
		    \raisebox{-0.5\height}{\includegraphics[width=\sresize]{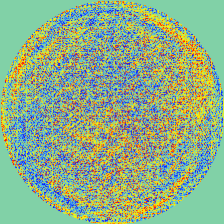} \includegraphics[width=\srescsize]{figures/simulated_results/colorbar}}
		    \smallskip \\
		    \midrule
		    \multirow{3}{*}[-44pt]{\rotatebox[origin=c]{90}{VGG16}} & &
		    \raisebox{-0.5\height}{\includegraphics[width=\sresize]{figures/simulated_results/starfish_noattack}} &
		    \STAB{\texttt{starfish} \\ (23.5\%)} & &
		    \raisebox{-0.5\height}{\includegraphics[width=\sresize]{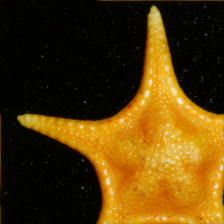}} &
		    \STAB{\texttt{\color{red}{flatworm}} \\ (18.5\%)} & &
		    \raisebox{-0.5\height}{\includegraphics[width=\sresize]{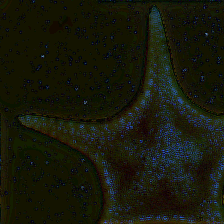}} &
		    \STAB{39.86 dB \\ 0.8799} & &
		    \raisebox{-0.5\height}{\includegraphics[width=\sresize]{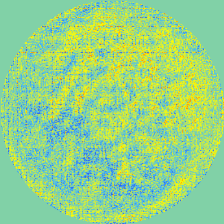} \includegraphics[width=\srescsize]{figures/simulated_results/colorbar}}
		    \smallskip \\
		    & &
		    \raisebox{-0.5\height}{\includegraphics[width=\sresize]{figures/simulated_results/butterfly_noattack}} &
		    \STAB{\texttt{cabbage} \\ \texttt{butterfly} \\ (71.6\%)} & &
		    \raisebox{-0.5\height}{\includegraphics[width=\sresize]{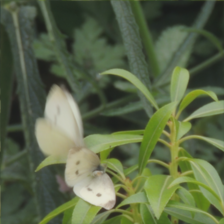}} &
		    \STAB{\texttt{\color{red}{sulphur}} \\ \texttt{\color{red}{butterfly}} \\ (45.1\%)} & &
		    \raisebox{-0.5\height}{\includegraphics[width=\sresize]{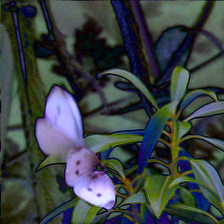}} &
		    \STAB{29.86 dB \\ 0.9803} & &
		    \raisebox{-0.5\height}{\includegraphics[width=\sresize]{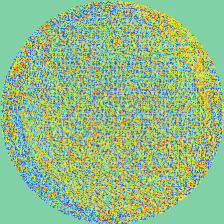} \includegraphics[width=\srescsize]{figures/simulated_results/colorbar}}
		    \smallskip \\
		    & &
		    \raisebox{-0.5\height}{\includegraphics[width=\sresize]{figures/simulated_results/airplane_noattack}} &
		    \STAB{\texttt{airliner} \\ (94.8\%)} & &
		    \raisebox{-0.5\height}{\includegraphics[width=\sresize]{figures/simulated_results/airplane_vgg16_img}} &
		    \STAB{\texttt{\color{red}{airship}} \\ (48.1\%)} & &
		    \raisebox{-0.5\height}{\includegraphics[width=\sresize]{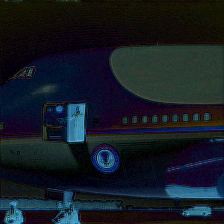}} &
		    \STAB{36.21 dB \\ 0.9952} & &
		    \raisebox{-0.5\height}{\includegraphics[width=\sresize]{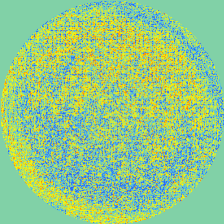} \includegraphics[width=\srescsize]{figures/simulated_results/colorbar}}
		    \smallskip \\
		    \midrule
		    \multirow{3}{*}[-32pt]{\rotatebox[origin=c]{90}{MobileNetV3}} & &
		    \raisebox{-0.5\height}{\includegraphics[width=\sresize]{figures/simulated_results/starfish_noattack}} &
		    \STAB{\texttt{starfish} \\ (98.2\%)} & &
		    \raisebox{-0.5\height}{\includegraphics[width=\sresize]{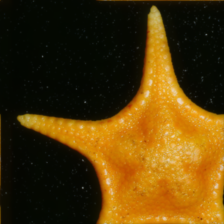}} &
		    \STAB{\texttt{\color{red}{mask}} \\ (46.4\%)} & &
		    \raisebox{-0.5\height}{\includegraphics[width=\sresize]{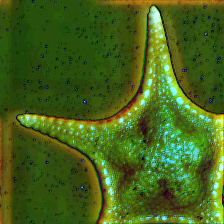}} &
		    \STAB{29.73 dB \\ 0.7004} & &
		    \raisebox{-0.5\height}{\includegraphics[width=\sresize]{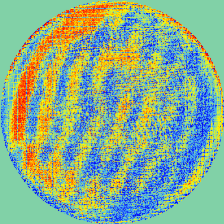} \includegraphics[width=\srescsize]{figures/simulated_results/colorbar}}
		    \smallskip \\
		    & &
		    \raisebox{-0.5\height}{\includegraphics[width=\sresize]{figures/simulated_results/butterfly_noattack}} &
		    \STAB{\texttt{cabbage} \\ \texttt{butterfly} \\ (72.3\%)} & &
		    \raisebox{-0.5\height}{\includegraphics[width=\sresize]{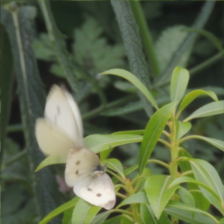}} &
		    \STAB{\texttt{\color{red}{axolotl}} \\ (33.0\%)} & &
		    \raisebox{-0.5\height}{\includegraphics[width=\sresize]{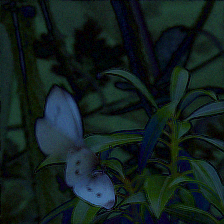}} &
		    \STAB{37.33 dB \\ 0.9953} & &
		    \raisebox{-0.5\height}{\includegraphics[width=\sresize]{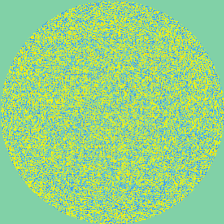} \includegraphics[width=\srescsize]{figures/simulated_results/colorbar}}
		    \smallskip \\
		    & &
		    \raisebox{-0.5\height}{\includegraphics[width=\sresize]{figures/simulated_results/airplane_noattack}} &
		    \STAB{\texttt{airliner} \\ (75.9\%)} & &
		    \raisebox{-0.5\height}{\includegraphics[width=\sresize]{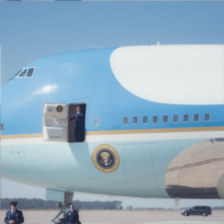}} &
		    \STAB{\texttt{\color{red}{airship}} \\ (49.0\%)} & &
		    \raisebox{-0.5\height}{\includegraphics[width=\sresize]{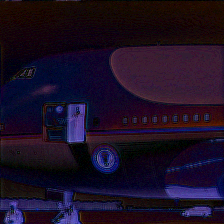}} &
		    \STAB{36.87 dB \\ 0.9960} & &
		    \raisebox{-0.5\height}{\includegraphics[width=\sresize]{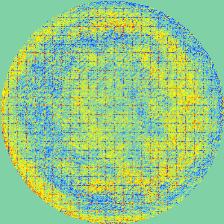} \includegraphics[width=\srescsize]{figures/simulated_results/colorbar}}
		    \smallskip
		\end{tabular}
	\end{center}
	\caption{Visual showcases of the unattacked and attacked images for the three image classification models. Classified labels and their confidence levels are also reported. The third column shows the absolute pixel value differences between the unattacked and attacked images, which are magnified 10 times for better visualization. The last column shows the modulated phase in the Fourier domain.}
	\label{fig:simulated_results_example}
\end{figure}

Figure~\ref{fig:simulated_results_example} shows example images with and without the adversarial attack.
The absolute differences of the unattacked and attacked images in the digital domain and the optimized phase modulation patterns ($\phi$ in Section~\ref{subsec:Imaging_model}) are also shown.
It can be seen that differences between the original and attacked images are not significant, which is also shown as high PSNR and SSIM values.
However, the classification models misclassify all the attacked images.
Here, the classified labels differ depending on the employed models.
For instance, the \texttt{starfish} image is misclassified as \texttt{honeycomb}, \texttt{flatworm}, and \texttt{mask} by each model, respectively.
The pixel-domain changes also differ depending on the employed models.
For example, the differences are mostly on the red channel for ResNet50, while those are mostly on the green channel for MobileNetV3.
The amount of distortion is also model-dependent, i.e., the PSNR and SSIM values differ depending on the target classification model for the same image.
For example, the PSNR values of the \texttt{cabbage butterfly} image for VGG16 and MobileNetV3 are 29.86 dB and 37.33 dB, respectively.
These model-dependent characteristics of the perturbations can be also found from low transferability of the attacked images between the models as shown in Table~\ref{table:simulated_transferability}.

\newlength{\oldintextsep}
\setlength{\oldintextsep}{\intextsep}
\setlength{\intextsep}{0pt}
\begin{wraptable}{r}{0.48\linewidth}
    \vspace{-18pt}
    \caption{Transferability of attacked images for different models in terms of accuracy}
    \label{table:simulated_transferability}
    \centering
    \renewcommand{\tabcolsep}{2.0pt}
    \vspace{6pt}
    \footnotesize
    \begin{tabular}{c|ccc}
        \toprule
        ~\backslashbox{Source}{Target}~ & ResNet50 & VGG16 & MobileNetV3 \\
        \midrule
        ResNet50 & 0.334 & 0.792 & 0.840 \\
        VGG16 & 0.851 & 0.260 & 0.823 \\
        MobileNetV3 & 0.862 & 0.782 & 0.323 \\
        \bottomrule
    \end{tabular}
    \vspace{18pt}
\end{wraptable}
\setlength{\intextsep}{\oldintextsep}

However, we also observe the following characteristics of the phase modulation patterns for different images and classifiers.
First, a wider range of phase modulations tends to yield a more distorted image having a lower PSNR value.
For instance, for ResNet50, the phase patterns of both the \texttt{starfish} and \texttt{airliner} images contain larger values (appearing as more red and blue colors) than that of the \texttt{cabbage butterfly} image.
Second, the phase patterns of the same image appear similar to some extent across different models.
For example, the phases of \texttt{starfish} show wave-like patterns, while those of \texttt{cabbage butterfly} contain more grain-like textures.

The overall patterns of the pixel value changes are largely different from those obtained from many existing adversarial attacks in the pixel domain \cite{goodfellow2014explaining,kurakin2016adversarial,carlini2017towards}.
The former preserves textures of the original images, whereas the latter is typically similar to random noise and barely preserves the original textures.
It is because our adversarial attack method manipulates the imaging system in the phase domain instead of the pixel domain.

\section{Real experiments}
\label{sec:exp_real}

We physically implement our proposed adversarial attack with an optical system as explained in Section~\ref{sec:system} in order to demonstrate the vulnerability of real optical systems in the wild to the proposed optical attack.

\subsection{Implementation details}

In the real experiments, we place ``actual'' objects in front of the optical system to capture and acquire images of the objects. 
Considering this practical constraint, we use ten real objects that correspond to ten ImageNet classes to obtain images of those objects in the digital domain, which are \texttt{bath towel}, \texttt{computer keyboard}, \texttt{lighter}, \texttt{paintbrush}, \texttt{ping-pong ball}, \texttt{plate rack}, \texttt{ruler}, \texttt{screwdriver}, \texttt{syringe}, and \texttt{toilet tissue}.
We place an object 100-120 cm away from the camera lens, which is a distance with a field-of-view of about $250 \times 250$ mm.
Phase modulation is performed using the SLM with a resolution of $224 \times 224$ pixels and a pixel size of 30 $\mathrm{\mu m}$.

We employ the pre-trained ResNet50 and VGG16 models.
The MobileNetV3 model is excluded here due to its relatively poor performance on the actual objects.

In addition to our attack method, we also investigate the impact of other optical-domain distortions that are usually found in real optical systems.
In this study, we consider spherical aberration, defocus, and astigmatism.
The amounts of these distortions are determined in a way that the resulting images have similar SSIM values to those perturbed by our attack.

\subsection{Results}

\begin{table}
    \caption{Performance comparison in terms of accuracy, PSNR, and SSIM for the original images, attacked images in simulation, attacked images in the real system, and images with optical distortions in the real system. Standard deviations across images are also shown.}
    \label{table:real_results_accuracy}
    \centering
    \renewcommand{\tabcolsep}{4.5pt}
    \newcommand{\STABL}[1]{{\renewcommand{\arraystretch}{0.7}{\STAB{\\[-6pt] #1 \\[1pt]}}}}
    \begin{tabular}{c|c|cccccc}
        \toprule
        Model & Metric & Original & Simulation & Real & Aberration & Defocus & Astigmatism \\
        \midrule
        \multirow{4.6}{*}{ResNet50} & Accuracy & 1.0 & 0.0 & 0.0 & 1.0 & 1.0 & 1.0 \\
        & PSNR (dB) & -- & -- & \STABL{33.36 \\ {\scriptsize{($ \pm 3.53 $)}}} & \STABL{35.81 \\ {\scriptsize{($ \pm 1.83 $)}}} & \STABL{36.03 \\ {\scriptsize{($ \pm 2.08 $)}}} & \STABL{36.11 \\ {\scriptsize{($ \pm 2.12 $)}}} \\
        & SSIM & -- & -- & \STABL{0.9357 \\ {\scriptsize{($ \pm 0.0263 $)}}} & \STABL{0.9391 \\ {\scriptsize{($ \pm 0.0191 $)}}} & \STABL{0.9386 \\ {\scriptsize{($ \pm 0.0195 $)}}} & \STABL{0.9394 \\ {\scriptsize{($ \pm 0.0206 $)}}} \\
        \midrule
        \multirow{4.6}{*}{VGG16} & Accuracy & 1.0 & 0.0 & 0.0 & 0.9 & 0.9 & 0.9 \\
        & PSNR (dB) & -- & -- & \STABL{34.31 \\ {\scriptsize{($ \pm 2.75 $)}}} & \STABL{35.38 \\ {\scriptsize{($ \pm 1.75 $)}}} & \STABL{35.37 \\ {\scriptsize{($ \pm 2.11 $)}}} & \STABL{35.29 \\ {\scriptsize{($ \pm 2.49 $)}}} \\
        & SSIM & -- & -- & \STABL{0.9381 \\ {\scriptsize{($ \pm 0.0323 $)}}} & \STABL{0.9380 \\ {\scriptsize{($ \pm 0.0170 $)}}} & \STABL{0.9366 \\ {\scriptsize{($ \pm 0.0191 $)}}} & \STABL{0.9367 \\ {\scriptsize{($ \pm 0.0212 $)}}} \\
        \bottomrule
    \end{tabular}
\end{table}

Table~\ref{table:real_results_accuracy} shows the overall performance comparison of our attack method and the three optical distortions for different image classification models, where the accuracy, PSNR, and SSIM values are reported.
We also report the accuracy of the original images and the attacked images obtained from our simulation environment explained in Section~\ref{sec:exp_simulation}.
Both ResNet50 and VGG16 successfully classify the ten real objects when no distortion is involved.
However, when our attack method is employed, all the objects are classified incorrectly for both models in both simulation and real environments.
Furthermore, all optical-domain distortions do not affect much the classification performance unlike our attack method; all ten objects are still classified correctly for ResNet50 and nine objects for VGG16.
These results demonstrate that the real optical system is highly vulnerable to the proposed optical adversarial attack. 

\begin{figure} 
	\begin{center}
		\centering
		\renewcommand{\tabcolsep}{2.0pt}
		\newcommand{\rresize}{0.1415\linewidth}
		\newcommand{\rrescsize}{0.20\linewidth}
		\footnotesize
		\begin{tabular}{cccccccccc}
		    & & \textbf{Original} & & \multicolumn{2}{c}{\textbf{Attacked}} & & \multicolumn{3}{c}{\textbf{Optical distortions}} \\
		    & ~ & & ~ & Simulation & Real & ~ & Aberration & Defocus & Astigmatism \\
		    \midrule
		    \#1 & &
		    \scriptsize
		    \STAB{\includegraphics[width=\rresize]{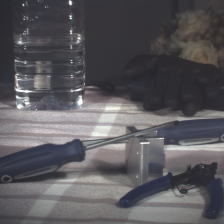} \\ \makebox[\rresize]{\rule{0pt}{\rresize}} \\ \texttt{screwdriver} \\ (42.7\%)} &
		    &
		    \scriptsize
		    \STAB{\includegraphics[width=\rresize]{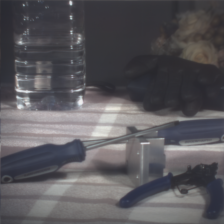} \\ \includegraphics[width=\rresize]{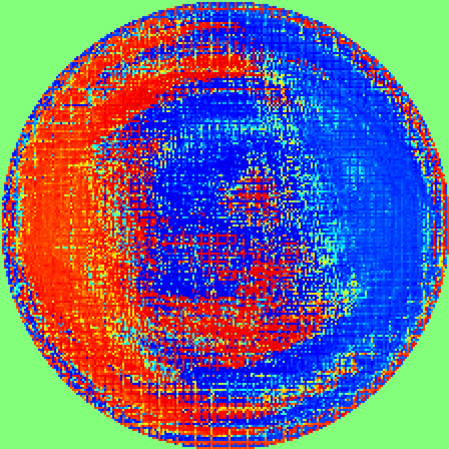} \\ \texttt{\color{red}{assault rifle}} \\ (18.5\%)} &
		    \scriptsize
		    \STAB{\includegraphics[width=\rresize]{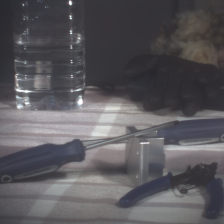} \\ \includegraphics[width=\rresize]{figures/real_results_resnet50/driver/res_pattern} \\ \texttt{\color{red}{assault rifle}} \\ (17.9\%)} &
		    &
		    \scriptsize
		    \STAB{\includegraphics[width=\rresize]{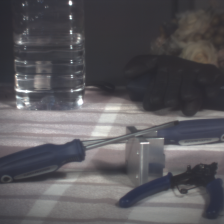} \\ \includegraphics[width=\rresize]{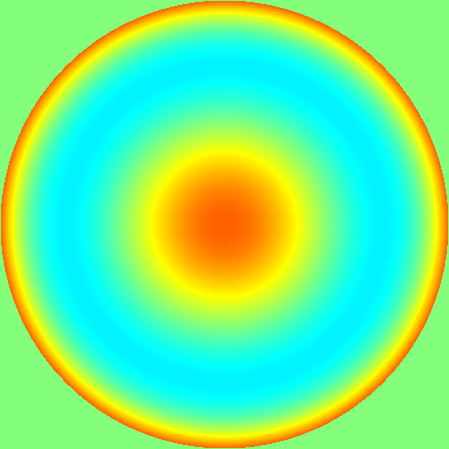} \\ \texttt{screwdriver} \\ (21.1\%)} &
		    \scriptsize
		    \STAB{\includegraphics[width=\rresize]{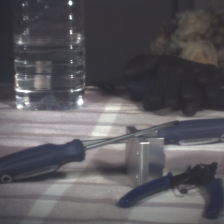} \\ \includegraphics[width=\rresize]{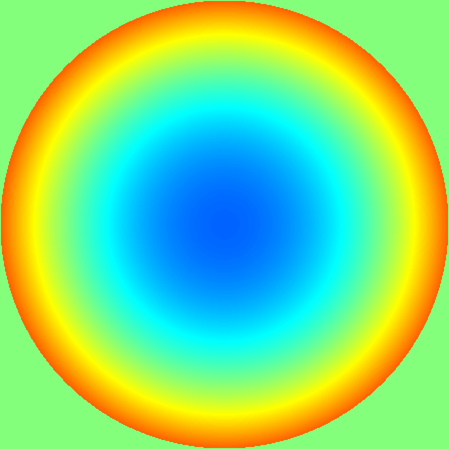} \\ \texttt{screwdriver} \\ (23.1\%)} &
		    \scriptsize
		    \STAB{\includegraphics[width=\rresize]{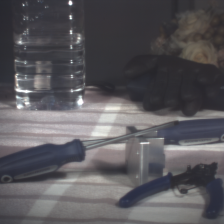} \\ \includegraphics[width=\rresize]{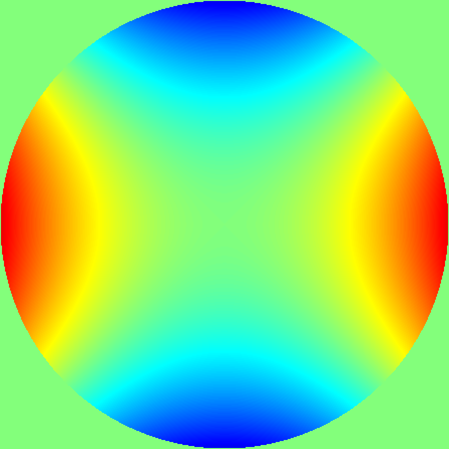} \\ \texttt{screwdriver} \\ (21.5\%)}
		    \smallskip \\
		    \midrule
		    \#2 & &
		    \scriptsize
		    \STAB{\includegraphics[width=\rresize]{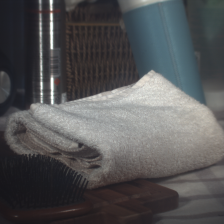} \\ \makebox[\rresize]{\rule{0pt}{\rresize}} \\ \texttt{bath towel} \\ (29.5\%)} &
		    &
		    \scriptsize
		    \STAB{\includegraphics[width=\rresize]{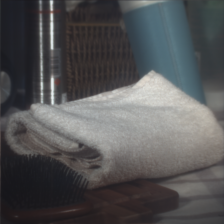} \\ \includegraphics[width=\rresize]{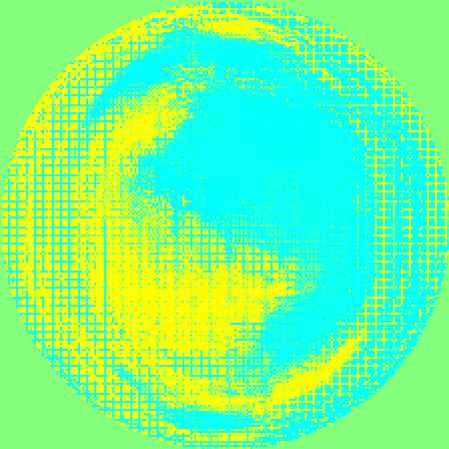} \\ \texttt{\color{red}{paper towel}} \\ (24.0\%)} &
		    \scriptsize
		    \STAB{\includegraphics[width=\rresize]{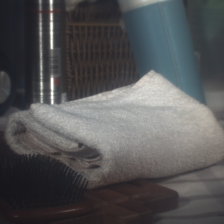} \\ \includegraphics[width=\rresize]{figures/real_results_resnet50/towel/res_pattern} \\ \texttt{\color{red}{paper towel}} \\ (23.6\%)} &
		    &
		    \scriptsize
		    \STAB{\includegraphics[width=\rresize]{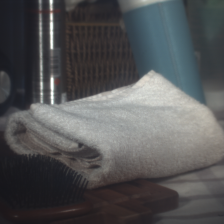} \\ \includegraphics[width=\rresize]{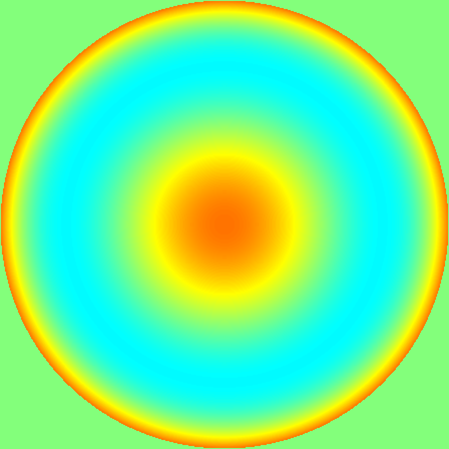} \\ \texttt{bath towel} \\ (18.9\%)} &
		    \scriptsize
		    \STAB{\includegraphics[width=\rresize]{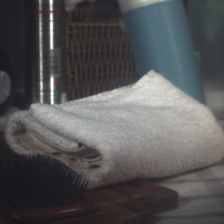} \\ \includegraphics[width=\rresize]{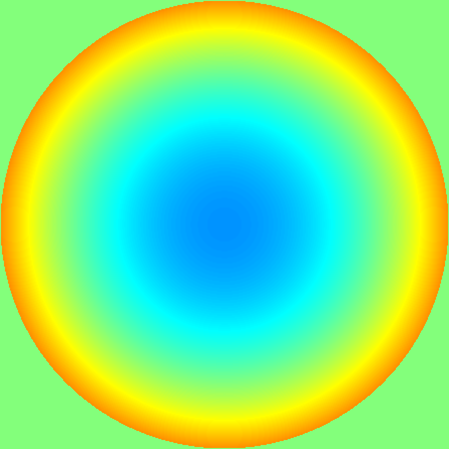} \\ \texttt{bath towel} \\ (16.9\%)} &
		    \scriptsize
		    \STAB{\includegraphics[width=\rresize]{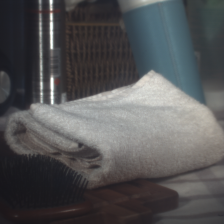} \\ \includegraphics[width=\rresize]{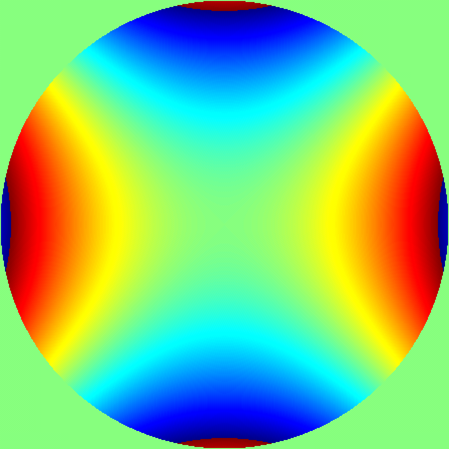} \\ \texttt{bath towel} \\ (14.2\%)}
		    \smallskip \\
		    \midrule
		    \#3 & &
		    \scriptsize
		    \STAB{\includegraphics[width=\rresize]{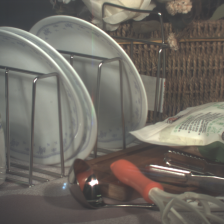} \\ \makebox[\rresize]{\rule{0pt}{\rresize}} \\ \texttt{plate rack} \\ (20.7\%)} &
		    &
		    \scriptsize
		    \STAB{\includegraphics[width=\rresize]{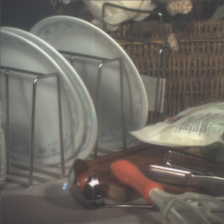} \\ \includegraphics[width=\rresize]{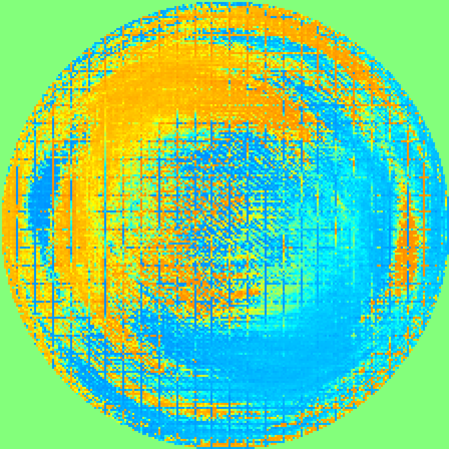} \\ \texttt{\color{red}{minivan}} \\ (12.4\%)} &
		    \scriptsize
		    \STAB{\includegraphics[width=\rresize]{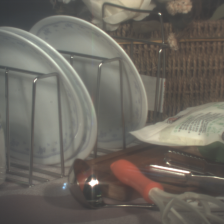} \\ \includegraphics[width=\rresize]{figures/real_results_resnet50/plate_rack/res_pattern} \\ \texttt{\color{red}{oxygen mask}} \\ (17.7\%)} &
		    &
		    \scriptsize
		    \STAB{\includegraphics[width=\rresize]{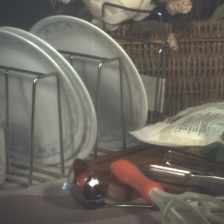} \\ \includegraphics[width=\rresize]{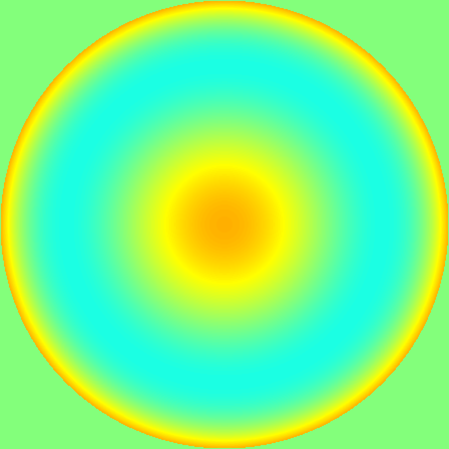} \\ \texttt{plate rack} \\ (14.8\%)} &
		    \scriptsize
		    \STAB{\includegraphics[width=\rresize]{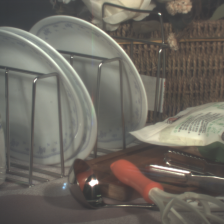} \\ \includegraphics[width=\rresize]{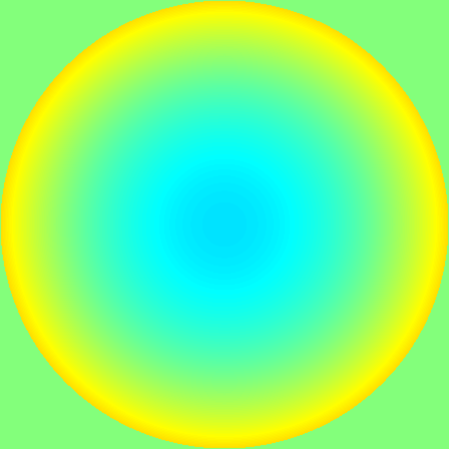} \\ \texttt{plate rack} \\ (14.9\%)} &
		    \scriptsize
		    \STAB{\includegraphics[width=\rresize]{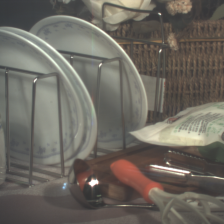} \\ \includegraphics[width=\rresize]{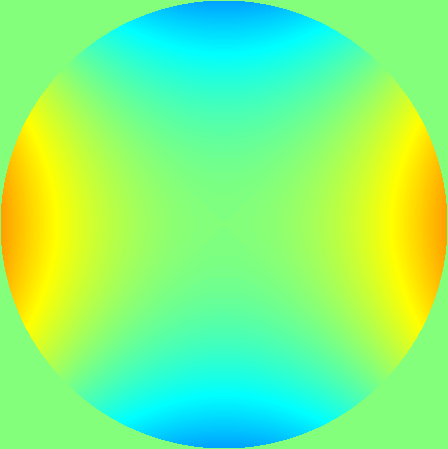} \\ \texttt{plate rack} \\ (17.9\%)}
		    \smallskip \\
		    \midrule
		    \#4 & &
		    \scriptsize
		    \STAB{\includegraphics[width=\rresize]{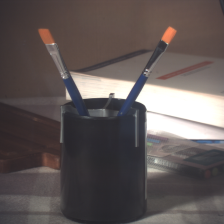} \\ \makebox[\rresize]{\rule{0pt}{\rresize}} \\ \texttt{paintbrush} \\ (30.7\%)} &
		    &
		    \scriptsize
		    \STAB{\includegraphics[width=\rresize]{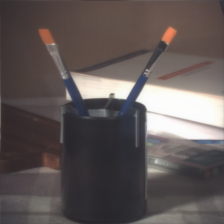} \\ \includegraphics[width=\rresize]{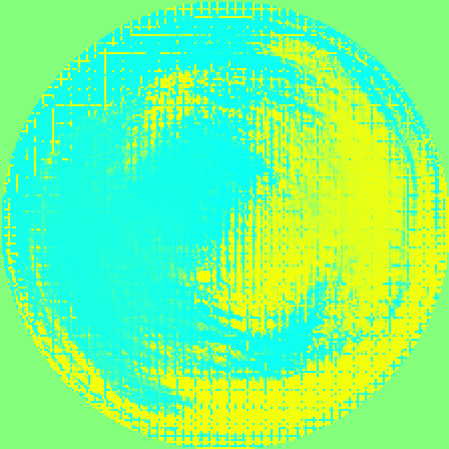} \\ \texttt{\color{red}{mortar}} \\ (25.4\%)} &
		    \scriptsize
		    \STAB{\includegraphics[width=\rresize]{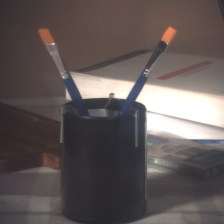} \\ \includegraphics[width=\rresize]{figures/real_results_resnet50/paint_brush/res_pattern} \\ \texttt{\color{red}{mortar}} \\ (26.5\%)} &
		    &
		    \scriptsize
		    \STAB{\includegraphics[width=\rresize]{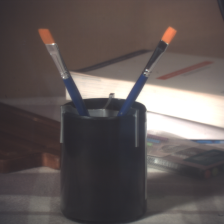} \\ \includegraphics[width=\rresize]{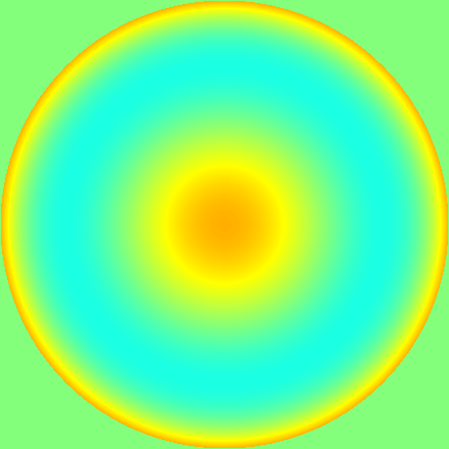} \\ \texttt{paintbrush} \\ (26.6\%)} &
		    \scriptsize
		    \STAB{\includegraphics[width=\rresize]{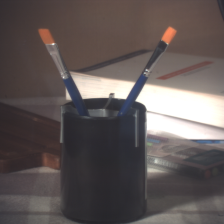} \\ \includegraphics[width=\rresize]{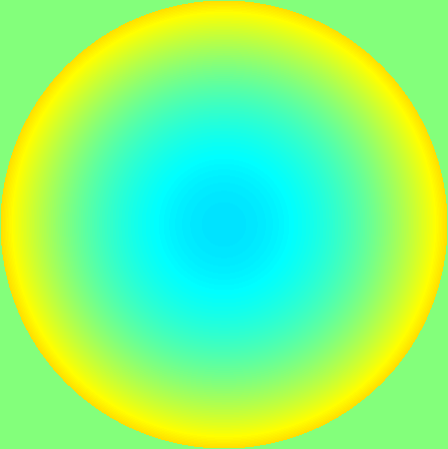} \\ \texttt{paintbrush} \\ (26.9\%)} &
		    \scriptsize
		    \STAB{\includegraphics[width=\rresize]{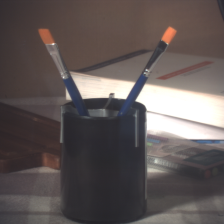} \\ \includegraphics[width=\rresize]{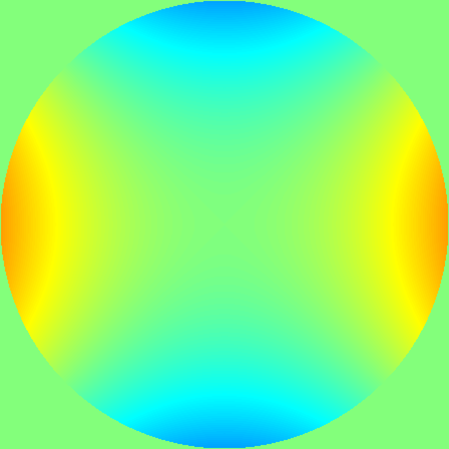} \\ \texttt{paintbrush} \\ (27.3\%)}
		    \smallskip \\
		    & & & &
		    \multicolumn{2}{c}{\includegraphics[width=\rrescsize]{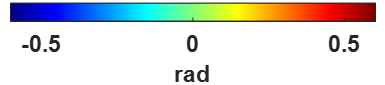}} &
		    &
		    \multicolumn{3}{c}{\includegraphics[width=\rrescsize]{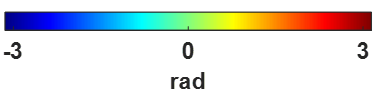}}
		    \smallskip
		\end{tabular}
	\end{center}
	\caption{Visual showcases of the original and attacked images for ResNet50. Classified labels and their confidence levels are also reported.}
	\label{fig:real_results_example_resnet50}
\end{figure}

Figure~\ref{fig:real_results_example_resnet50} shows four visual showcases of our attack and the optical distortions for ResNet50.
When the original and attacked outputs are compared in the digital domain, there are no obvious visual differences between them, which also appears as high PSNR in Table~\ref{table:real_results_accuracy}.
However, our attack method successfully fools the target image classification model.
For example, the original image \#4 is correctly classified as \texttt{paintbrush}.
However, the attacked ones are misclassified as \texttt{mortar} in the simulation and real environments.
The optical distortions hardly affect the classification performance, reducing the confidence levels only slightly (e.g., from 30.7\% to 26.6\% by spherical aberration) without resulting in misclassification.

The patterns of the phase images obtained from our attack method and those of the optical distortions also show significant differences.
First, the range of the phase values for our attack is significantly smaller that those for the optical distortions: about $\left[ -0.5, 0.5 \right]$ (rad) vs. $\left[ -3, 3 \right]$ (rad).
In addition, the phase patterns are highly distinguishable across different objects for our attack method, while they are not for the optical distortions.

\section{Conclusion}
\label{sec:conclusion}

We presented the feasibility of attacking optical systems in the optical domain instead of attacking images in the digital domain by introducing an optical adversarial attack.
For a given real object, our attack method finds a spatially varying phase modulation pattern implemented by an SLM in order to minimize the amount of distortion in the digital domain but significantly degrade the performance of image classification models.
We conducted experiments not only in a simulation environment to evaluate with a large amount of data but also in a real optical system to evaluate the proposed attack method in the wild.
The results showed that the optical systems are highly vulnerable to our adversarial attack method, raising a new significant security issue of imaging systems.

Our current work has the following limitations that call for future work.
First, our study can be expanded to a broader range of application fields.
Although we considered the image classification task as the main target of our experiments, our proposed adversarial attack method can be further applied to other fields that employ optical imaging systems to acquire digital images and deep neural networks to perform classification or enhancement, such as microscopic image enhancement \cite{rivenson17deep} and hologram classification \cite{kim18deep}.
Second, we focused on investigating the vulnerability of the physical imaging system in this study, and to this end, we proposed the optical adversarial attack.
One of the important next directions will be to find ways to protect optical systems against adversarial attacks applied in both the optical domain and the digital domain.

{\small
\bibliographystyle{unsrt}
\bibliography{optical_adversarial_attack}
}

\end{document}